\documentclass[runningheads]{llncs}
\usepackage[T1]{fontenc}
\usepackage{xcolor}
\usepackage{hyperref}
\usepackage{algorithm} 
\usepackage{algpseudocode} 
\usepackage{cite}
\usepackage{amsmath,amssymb,amsfonts}
\usepackage{graphicx}
\usepackage{textcomp}
\usepackage{xcolor}
\usepackage{hyperref}
\usepackage{fancyhdr}
\usepackage{graphicx}
\usepackage{pifont}
\newcommand{\cmark}{\textcolor{blue}{\ding{51}}}%
\newcommand{\xmark}{\textcolor{red}{\ding{55}}}%

\algnewcommand\algorithmicinput{\textbf{Input:}}
\algnewcommand\Input{\item[\algorithmicinput]}

\algnewcommand\algorithmicoutput{\textbf{Output:}}
\algnewcommand\Output{\item[\algorithmicoutput]}

\fancypagestyle{firstpage}
{
    \fancyhead[C]{To appear at the 18th International Symposium on Visual Computing (ISVC’23), October 2023, Lake Tahoe, NV, USA.}    
    \fancyhead[R]{}
}

\begin{document}
\title{ISLE: A Framework for Image Level Semantic Segmentation Ensemble}
%
%\titlerunning{Abbreviated paper title}
% If the paper title is too long for the running head, you can set
% an abbreviated paper title here
%
\author{Erik Ostrowski\inst{1}%\orcidID{1111-2222-3333-4444}
\and
Muhammad Shafique\inst{2}}%\orcidID{2222--3333-4444-5555}}

%\author{
%    \IEEEauthorblockN{
%    Erik Ostrowski\IEEEauthorrefmark{1}, Bharath Srinivas Prabakaran\IEEEauthorrefmark{1}, Muhammad Shafique\IEEEauthorrefmark{3}}
%    \IEEEauthorblockA{\IEEEauthorrefmark{1}Institute of Computer Engineering, Technische Universit{\"a}t Wien (TU Wien), Austria
%    \\ \{erik.ostrowski, bharath.prabakaran\}@tuwien.ac.at}
%    \IEEEauthorblockA{\IEEEauthorrefmark{3}Division of Engineering, New York University Abu Dhabi (NYUAD), United Arab Emirates (UAE)
%    \\muhammad.shafique@nyu.edu}
%}
\authorrunning{ }
% First names are abbreviated in the running head.
% If there are more than two authors, 'et al.' is used.
%
\institute{Institute of Computer Engineering, Technische Universit{\"a}t Wien (TU Wien), Austria\\
\email{erik.ostrowski@tuwien.ac.at}
\and
eBrain Lab, Division of Engineering, New York University Abu Dhabi (NYUAD), United Arab Emirates (UAE)\\
\email{muhammad.shafique@nyu.edu}}

\maketitle              % typeset the header of the contribution
\thispagestyle{firstpage}

\begin{abstract}
One key bottleneck of employing state-of-the-art semantic segmentation networks in the real world is the availability of training labels.
Conventional semantic segmentation networks require massive pixel-wise annotated labels to reach state-of-the-art prediction quality.
Hence, several works focus on semantic segmentation networks trained with only image-level annotations.
However, when scrutinizing the results of state-of-the-art in more detail, we notice that they are remarkably close to each other on average prediction quality, different approaches perform better in different classes while providing low quality in others.
To address this problem, we propose a novel framework, ISLE, which employs an ensemble of the "pseudo-labels" for a given set of different semantic segmentation techniques on a class-wise level.
Pseudo-labels are the pixel-wise predictions of the image-level semantic segmentation frameworks used to train the final segmentation model.
Our pseudo-labels seamlessly combine the strong points of multiple segmentation techniques approaches to reach superior prediction quality.
We reach up to 2.4\% improvement over ISLE's individual components.
An exhaustive analysis was performed to demonstrate ISLE's effectiveness over state-of-the-art frameworks for image-level semantic segmentation.
\end{abstract}

\begin{keywords}
Semantic Segmentation, Weakly Supervised, Ensemble, Deep Learning, Class Activation Maps
\end{keywords}
%%%%%%%%%%%%%%%%%%%%%%%%%%%%%%%%%%%%%%%%%%%%%%%%%%%%%%%%%%%%%%%%%%%%%%%%%%%%%%%%%%%%%%%%%%%%%%%%%%%%%%%%%%%%%%%%
% 1. SECTION INTRODUCTION
%%%%%%%%%%%%%%%%%%%%%%%%%%%%%%%%%%%%%%%%%%%%%%%%%%%%%%%%%%%%%%%%%%%%%%%%%%%%%%%%%%%%%%%%%%%%%%%%%%%%%%%%%%%%%%%%

\section{Introduction}

\begin{figure}[t]
\centering
\includegraphics[width=
100mm,scale=0.8]{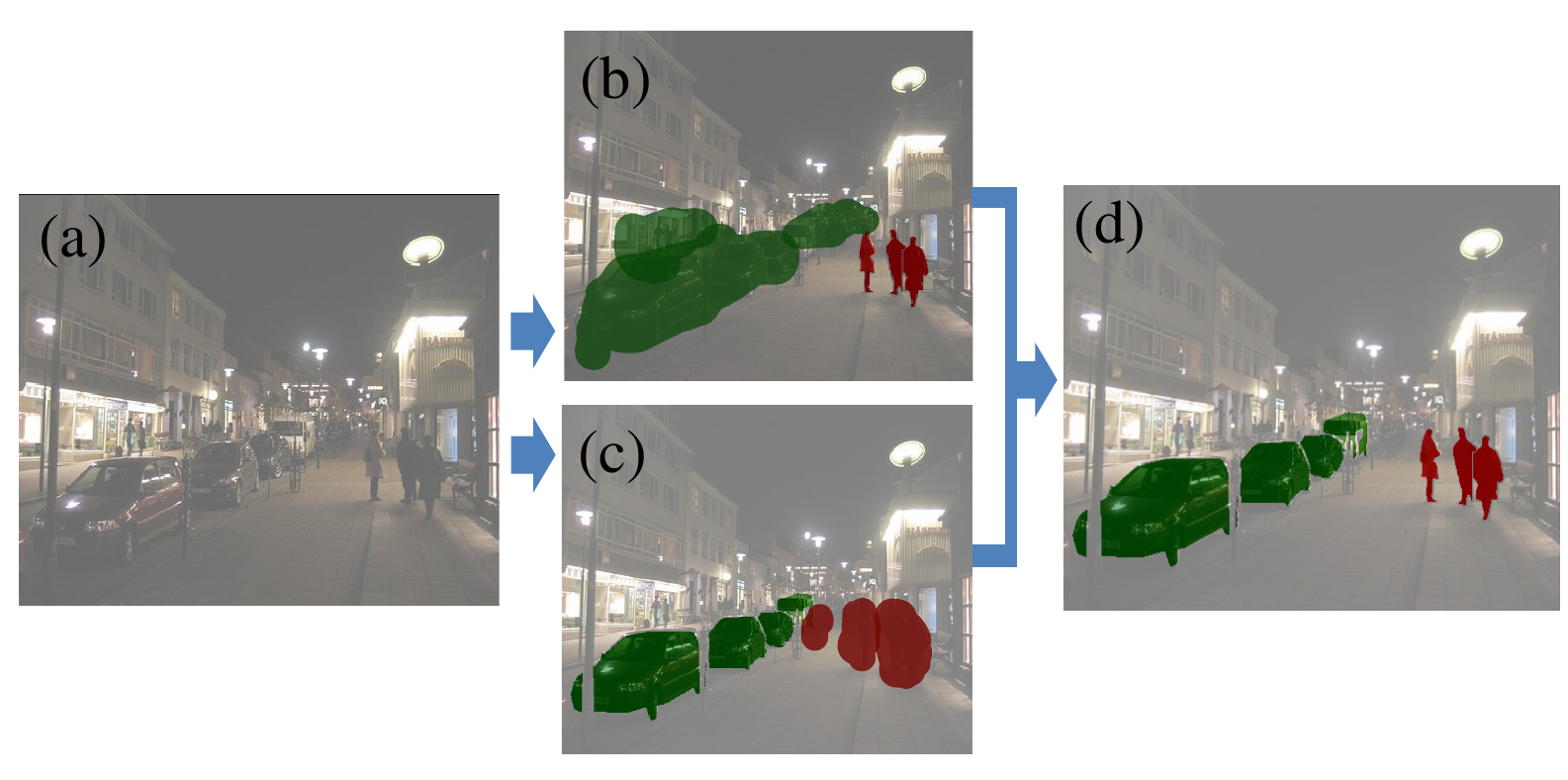}
\caption{The idea of our Framework: (a) Input image, (b) prediction of a method that is good with person segmentation and bad with cars, (c) prediction of a method good with car segmentation and bad with persons, (d) prediction of ISLE, combining the strengths of the two methods.\label{example}}

\end{figure}

Generating high-quality semantic segmentation predictions using only models trained on image-level annotations would enable a new level of applicability.
The progress of fully supervised semantic segmentation networks has already helped provide many useful tools and applications. For example, in autonomous and self-driving vehicles~\cite{AD,feng2020deep}, remote sensing~\cite{diakogiannis2020resunet}, facial% , kemker2018algorithms
recognition~\cite{meenpal2019facial,khan2015multi}, agriculture \cite{milioto2018real, barth2018data}, and in the medical field \cite{rehman2020deep,zhao2017tracking}, etc.
The downside of those fully supervised semantic segmentation networks (FSSS) is that they require copious amounts of pixel-wise annotated images. 
Generating such a training set is very tedious and time-consuming  work. 
For instance, one image of the Cityscapes dataset, which contains street scenes from cities that require many complex objects to be annotated, takes more than an hour of manual user-driven labour~\cite{CT}.
Furthermore, medical imaging and molecular biology fields require the knowledge of highly qualified individuals capable of interpreting and annotating the images.

Therefore, to reduce the time and resources required for generating pixel-wise masks, a wide range of research works focus on developing approaches that focus on weaker kinds of supervision.
In this work, we will focus on weak supervision in the form of image-level labels.
Image-level labels give the least amount of supervision for semantic segmentation but are the easiest to acquire.

Several works already focus on image-level semantic segmentation techniques, and they consistently reach better and better high scores.
Most works are based on Class Activation Maps (CAMs)~\cite{CAM}.
CAMs localize the object by training a DNN model with classification loss and then reusing the learned weights to highlight the image areas responsible for its classification decision.
Most image-level segmentation approaches aim to improve the CAM baseline by adding additional regularizations to the classification loss or refining the CAM mask afterward.
As more methods emerge for improving CAM quality, state-of-the-art is usually compiled of regularizations, after-the-fact refinements, or combinations of both.
However, when analyzing multiple image-level segmentation techniques on a class-by-class basis, we observed that the differences between them vary significantly on specific classes, although those methods generate predictions that reach comparable scores on average. 

Therefore, we are proposing our ISLE framework.
In our framework, we combine the pseudo-labels of multiple image-level segmentation techniques based on the respective class scores to generate a superset of pseudo-labels, combining the upsides of multiple different approaches.
Fig.~\ref{example} visualizes the gains possible by our ISLE framework in comparison to its components.
The detection of edges of objects is a major weakness in Image-level based semantic segmentation and many state-of-the-art methods tried to address this problem. 
We noticed that they achieve their goal to some degree.
Namely, in certain objects, a specific method achieves remarkable results, but also scores under average with different objects.
Hence, our ISLE framework addresses the problem of insufficient object border detection, by applying state-of-the-art only in scenarios, where their individual approach is suited best.
We perform extensive experiments on the PASCAL VOC2012 dataset~\cite{VOC} to prove the effectiveness of our framework in various experimental settings and compare them with a range of state-of-the-art techniques to illustrate the benefits of our approach.
The \textbf{key contributions} of this work are:

\begin{enumerate}
    \item Our novel ISLE framework improves the prediction quality of segmentation masks by combining state-of-the-art pseudo-labels on a class-by-class basis.
   \item Our ISLE framework is not limited by the number or approach of any image-level guided segmentation frameworks to combine their pseudo-labels. 
   Since the ISLE is only used for generating pseudo-labels, it will not add more computations for inference predictions.
   \item We present detailed ablation studies and analysis comparing the results of ISLE to state-of-the-art methods on the VOC2012 dataset to evaluate our method's efficacy and the improvements achieved using our framework. 
   \item The complete framework is open-source and accessible online at \url{https://github.com/ErikOstrowski/ISLE}.

\end{enumerate}

%%%%%%%%%%%%%%%%%%%%%%%%%%%%%%%%%%%%%%%%%%%%%%%%%%%%%%%%%%%%%%%%%%%%%%%%%%%%%%%%%%%%%%%%%%%%%%%%%%%%%%%%%%%%%%%%
% 2. SECTION RELATED WORK
%%%%%%%%%%%%%%%%%%%%%%%%%%%%%%%%%%%%%%%%%%%%%%%%%%%%%%%%%%%%%%%%%%%%%%%%%%%%%%%%%%%%%%%%%%%%%%%%%%%%%%%%%%%%%%%%

\section{Related Work}

In this section, we provide a discussion of the current state-of-the-art in semantic segmentation using image-level supervision.

AffinityNet~\cite{AFF} trains a second network to learn pixel similarities, which generates a transition matrix combined with the CAM iteratively to refine its activation coverage.
PuzzleCAM~\cite{PUZZLE} introduces a loss, that subdivides the input image into multiple parts, forcing the network to predict image segments that contain the non-discriminative parts of an object.
CLIMS~\cite{CLIMS} trained the network by matching text labels to the correct image.
Hence, the network maximizes and minimizes the distance between correct and wrong pairs, respectively, instead of just giving a binary classification result.
PMM~\cite{PMM} used Coefficient of Variation Smoothing to smooth the CAMs, which introduces a new metric, that highlights the importance of each class on each location, in contrast to the scores trained from the binary classifier.
Furthermore, they employed Pretended Under-fitting, which improves training with noisy labels, and Cyclic Pseudo-mask to iteratively trains the final segmentation network with its predictions.
DRS~\cite{DRS} aims to improve the image's activation area to less discriminative areas.
Kim et al. \cite{DRS} achieve this by suppressing the attention on discriminative regions, thus guiding the attention to adjacent regions to generate a complete attention map of the target object.

Table~\ref{check} lists all the twenty classes of the VOC2012 dataset and shows, which state-of-the-art method achieves the best result on specific classes.
We observe that PMM has the best result in only two of the twenty classes, DRS in three, CLIMS has the best result in seven, and PuzzleCAM in eight classes.

\begin{table}[ht]
\caption{Highest score per VOC2012 class on each component of ISLE.\label{check}}
\centering 
{
\begin{tabular}{p{0.9cm} p{0.8cm} p{0.8cm} p{1.0cm} p{0.8cm}}
\hline
Class&PMM&DRS&CLIMS&Puzzle\\ \hline
\begin{tabular}{p{0.9cm} p{0.8cm} p{0.8cm} p{1.0cm} p{0.8cm}} 
Bus    & \cmark  &   \xmark &   \xmark &   \xmark  \\
Car    & \cmark  &   \xmark &   \xmark &   \xmark  \\
Bottle & \xmark  &   \cmark &   \xmark &   \xmark  \\
Chair  & \xmark  &   \cmark &   \xmark &   \xmark  \\
Train  & \xmark  &   \cmark &   \xmark &   \xmark  \\
Bike   & \xmark  &   \xmark &   \cmark &   \xmark  \\
Boat   & \xmark  &   \xmark &   \cmark &   \xmark  \\
Table  & \xmark  &   \xmark &   \cmark &   \xmark  \\
Motor  & \xmark  &   \xmark &   \cmark &   \xmark  \\
Person & \xmark  &   \xmark &   \cmark &   \xmark  \\
Sofa   & \xmark  &   \xmark &   \cmark &   \xmark  \\
TV     & \xmark  &   \xmark &   \cmark &   \xmark  \\
Aero   & \xmark  &   \xmark &   \xmark &   \cmark   \\
Bird   & \xmark  &   \xmark &   \xmark &   \cmark   \\
Cat    & \xmark  &   \xmark &   \xmark &   \cmark   \\
Cow    & \xmark  &   \xmark &   \xmark &   \cmark   \\
Dog    & \xmark  &   \xmark &   \xmark &   \cmark  \\
Horse  & \xmark  &   \xmark &   \xmark &   \cmark   \\
Plant  & \xmark  &   \xmark &   \xmark &   \cmark   \\
Sheep  & \xmark  &   \xmark &   \xmark &   \cmark   \end{tabular} 
\end{tabular}}
\end{table}

%%%%%%%%%%%%%%%%%%%%%%%%%%%%%%%%%%%%%%%%%%%%%%%%%%%%%%%%%%%%%%%%%%%%%%%%%%%%%%%%%%%%%%%%%%%%%%%%%%%%%%%%%%%%%%%%
% 3. SECTION METHOD
%%%%%%%%%%%%%%%%%%%%%%%%%%%%%%%%%%%%%%%%%%%%%%%%%%%%%%%%%%%%%%%%%%%%%%%%%%%%%%%%%%%%%%%%%%%%%%%%%%%%%%%%%%%%%%%%

\section{ISLE Framework}

\begin{figure*}[ht]
\centering
\includegraphics[width=\columnwidth]{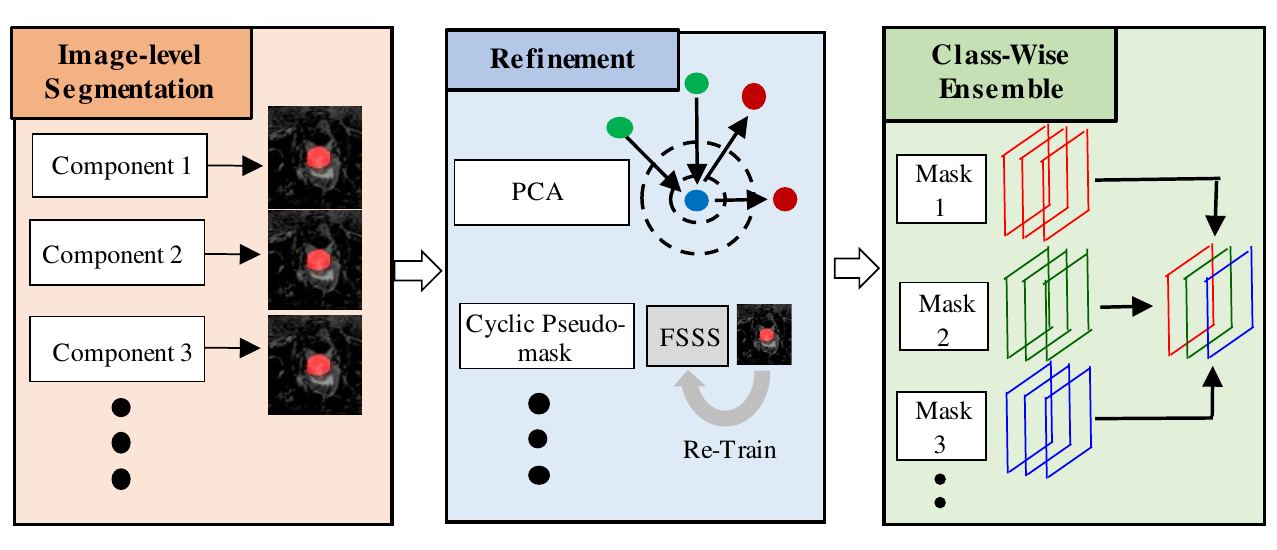} %width=80mm,scale=0.80 width=\linewidth
\caption{Overview of the ISLE framework. 
The first stage is the collection of Image-level semantic segmentation;
In the second stage, we can use a number of refinement methods to improve the mask quality; 
In the third stage, we combine the refinement masks on a class-wise basis to generate the pseudo-labels that reach the best prediction for each class;
In the final stage, we are training an FSSS with the pseudo labels.\label{fig2}}\end{figure*}

Our framework aims to combine the strengths of different methods by just using their predictions for classes where they are performing the best, while not considering predictions of classes where they perform worse compared to other methods.
The prerequisite of our framework is a list of candidate state-of-the-art approaches for image-level semantic segmentation.
Fig.~\ref{fig2} presents an overview of the ISLE framework.
We start with collecting the pseudo labels of our candidate methods.
In the next step, we can employ several refinement methods to improve the pseudo-label quality beforehand.
In our case, we used AffinityNet~\cite{AFF} and a dense Conditional Random Field (dCRF)~\cite{CRF} for the candidates if the provided pseudo labels did not already undergo refinement methods.
Then we can combine the pseudo-labels on a class-wise basis, where we only copy the predictions of classes of the candidate labels to our ensemble if the candidate has a high score in that class.
Finally, we use the generated pseudo labels to train an FSSS network.
Our proposed version uses the four state-of-the-art methods introduced in the previous section, and for all of them except CLIMS, we also refine their baseline with AffinityNet.
%%%%%%%%%%%%%%%%%%%%%%%%%%%%%%%%%%%%%%%%%%%%%%%%%%%%%%%%%%%%%%%%%%%%%%%%%%%%%%%%%%%%%%%%%%%%%%%%%%
AffinityNet uses a random walk, in combination with pixel affinities.
Furthermore, as a widespread practice, we use dCRF to the AffinityNet predictions to improve their quality.
The refinement of the pseudo labels is not limited to AffinityNet, any combination of additional refinement methods can be employed within our framework.

In the next step, we will evaluate the different candidate pseudo label sets on a class-wise basis and determine which candidate is used for which class for the ensemble.
We reshape the pseudo labels $ps$ to the dimensions $ps_i \in \mathbb{R}^{H \times W \times C}$, if they were not already provided in that shape, where $i$ is the specific image, $H$ is the image height, $W$ is the image width and $C$ is the number of classes.
We perform the combination by copying for all images the slides $ps^x_{i,c} \in \mathbb{R}^{H \times W \times 1}$  of a specific class $c$ by the candidate network $x$, if $x$ has the high score on $c$.
We do this for every class $c$ in $C$ and then simply concatenate all slides to get a complete prediction, which contains only the best predictions state-of-the-art can muster on a class-wise level:
$$ps^{x_1}_{i,{c_1}} \times ps^{x_2}_{i,{c_2}}  \times … \times ps^{x_M}_{i,{c_N}} =  ps^{*}_{i},$$
Where $N$ is the number of classes in the used dataset, $M$ the number of candidate methods and $ps^{*}_{i}$ is the ensemble prediction of image $i$.
\begin{algorithm}[H]
\scriptsize
\caption{ISLE Step 1 (Optional)}\label{alg:step1}
\begin{algorithmic}[1]
\Input{$N$ training-pseudo-segmentations of components: $ps$}
\Output{$N$  refined training-pseudo-segmentations of components: $ps^*$}\\

Let F() be a combination of refinement methods (F() = \{AffinityNet(), Cycle(),..\}.

\For {$n$ \textbf{in} range(N)}
\State $ps^*_n(i)=F(ps_n(i))$  
\State $ps_n(i) = ps^*_n(i)$
\EndFor
\end{algorithmic}
\end{algorithm}
Note that the same method can achieve the best score in multiple classes and therefore sometimes  $ x_i = x_j$ for $i \neq j$ for $i,j \in M$. 
Furthermore, we assessed a naive version, in which we ranked every class by its number of instances in the training set and then used the complete CAM of candidate $x$ from the whole image if $x$ has the high score on the highest ranked class $x$ present on the image.
The naive ISLE performed worse than our final version.

We excluded the \textit{background} class from our ensemble since the background is the inverse of all classes combined.
Note that we can perform this class selection method since we already assign the correct class labels to each prediction instead of using a classification network for the assignment, as conventional for image level based semantic segmentation methods.
Therefore, it is necessary to train a fully supervised semantic segmentation network with those pseudo-labels.
Nevertheless, the FSSS training guarantees that collecting multiple pseudo-label sets is a one-time effort per dataset.
Fig.~\ref{fig2} illustrates an overview of the process.
Further details can be seen in the Pseudocodes~\ref{alg:step1},~\ref{alg:step2},~\ref{alg:step3}.

%%%%%%%%%%%%%%%%%%%%%%%%%%%%%%%%%%%%%%%%%%%%%%%%%%%%%%%%%%%%%%%%%%%%%%%%%%%%%%%%%%%%%%%%%%%%%%%%%%%%%%%%%%%%%%%%

\begin{algorithm}[H]
\scriptsize
\caption{ISLE Step 2}\label{alg:step2}
\begin{algorithmic}[1]
\Input{N (refined) training-pseudo-segmentations of components: $ps$ }
\Output{List of best scoring methods for each class: best }\\
Let $C$ be the number of classes in the dataset.\\
Let $mIoU_c()$ be the evaluation algorithm on class $c$.\\
Let $Img$ be the list of all images.\\
\For {$c$ \textbf{in} range($C$)}
\State $top_c = 0$
\For {$n$ \textbf{in} range($N$)}
\For {$i$ \textbf{in} Img}

\State $score += mIoU_c(ps_n(i))$
\EndFor
\State $score = score / \# images$
\If { $score > top_c$}
\State $top_c = score$
\State $best(c)= n$
\EndIf 
\EndFor
\EndFor
\end{algorithmic}
\end{algorithm}

\begin{algorithm}[H]
\scriptsize
\caption{ISLE Step 3}\label{alg:step3}
\begin{algorithmic}[1]
\Input{$N$  refined training-pseudo-segmentations of components: $ps^*$ and list of best scoring methods for each class: $best$}
\Output{Semantic Segmentation network trained on weakly-supervised predictions: $SSN$}\\
Let $ae$ be the final ensemble.\\
Let $Img$ be the list of all images.\\
Define $ae = 0$ for all classes and all images in the dataset.\\ 
\For {$c$ \textbf{in} range($C$)}
\State $n = best(c)$
\For {$i$ \textbf{in} Img}
\State $ae_c(i) = ps_{n,c}(i)$
\EndFor
\EndFor\\ \\
\textbf{4.Step:}  \\
\For {Epochs $x$}
\For {$i$ \textbf{in} Img}
\State $pred = SSN(i)$
\EndFor
\State $Backpropagation(SSN)$  
\EndFor
\end{algorithmic}
\end{algorithm}

%%%%%%%%%%%%%%%%%%%%%%%%%%%%%%%%%%%%%%%%%%%%%%%%%%%%%%%%%%%%%%%%%%%%%%%%%%%%%%%%%%%%%%%%%%%%%%%%%%%%%%%%%%%%%%%%
% 4. SECTION EXPERIMENTS
%%%%%%%%%%%%%%%%%%%%%%%%%%%%%%%%%%%%%%%%%%%%%%%%%%%%%%%%%%%%%%%%%%%%%%%%%%%%%%%%%%%%%%%%%%%%%%%%%%%%%%%%%%%%%%%%

\section{Experiments}

First, we will discuss our experimental setup.
We completed the experiments on a CentOS 7.9 Operating System executing on an Intel Core i7-8700 CPU with 16GB RAM and 2 Nvidia GeForce GTX 1080 Ti GPUs.
The CLIMS pseudo-labels were used as provided by the official GitHub, and we performed AffinityNet with a ResNet50 backbone and dCRF on the pseudo-labels provided on the DRS and PMM GitHub.
All DeepLabV3+ results were generated using a ResNet50 backbone.
The mean Intersection-over-Union (mIoU) ratio is the evaluation metric for all experiments.
We used the PASCAL VOC2012 semantic segmentation benchmark for evaluating our framework. 
It comprises twenty-one classes, including a background class, and most images include multiple objects.
Following the conventional experimentation protocol for semantic segmentation, we use the $10,528$ augmented images and image-level labels, for training.
Our model is evaluated on the validation set with $1,464$ images and the test set of $1,456$ images to ensure a constant comparison with the state-of-the-art.

For all experiments, the mean Intersection-over-Union (mIoU) ratio is used as the evaluation metric.
%:
%Mean Average Precision (mAP) (AP [IoU=0.50:0.5:0.95], $AP^{50}$ [IoU=.50], $AP^{75}$[IoU=.75]) and mean Average Recall (mAR)(AR [IoU=0.50:0.5:0.95]).

%{%\fontsize{9pt}{9pt}\selectfont
%\begin{equation}
%\centering
%mIoU = \frac{1}{N} \sum_{i=1}^N \frac{ p_{i,i}}{\sum_{j=1}^N p_{i,j} + \sum_{j=1}^N p_{j,i} - %p_{i,i}  }
%\end{equation}
%}

%\noindent where $N$ is the total number of classes, $p_{i,i}$ the number of pixels classified as class $i$ when labelled as class $i$. $p_{i,j}$ and $p_{j,i}$ are the number of pixels classified as class $i$ that were labelled as class $j$ and vice-versa, respectively.

\begin{table}[ht]
%\resizebox{\columnwidth}{!}{
\centering 
{%\fontsize{9pt}{9pt}\selectfont
\caption{Comparison of ISLE mIoU scores with state-of-the-art techniques on the VOC2012 val and test datasets. All methods were trained with DeepLabV3+ with a ResNet50 backbone for comparability. Adding more or different methods to the ensemble is possible, as those work orthogonal to the ISLE.\label{tab2}}
\begin{tabular}{lccc}
\hline
Method              & Val  & Test \\ \hline
PuzzleCAM             & 62.4 & 62.9 \\
PMM            & 64.0 & 64.1 \\
DRS         & 64.5 &  64.5 \\
CLIMS             & 65.0 &  65.4\\
ISLE-2 (Ours)         & 66.6 & 67.1 \\ % 64.72
ISLE (Ours)         & \textbf{67.4} & \textbf{67.8} \\ \hline % 65.03
\end{tabular}}

\end{table}

\begin{table*}[ht]
%\resizebox{\textwidth}{!}{
\centering
\caption{Semantic segmentation mIoU performance on the first 11 classes of the VOC2012 \textbf{training dataset} for the final pseudo-labels. \label{tab3}}
{\begin{tabular}{lcccccccccccccccccccccc}

\hline
Method                              & bkg         & aero        & bike        & bird        & boat        & bottle      & bus         & car         & cat         & chair       & cow \\ \hline
%AffinityNet~\cite{AFF}                      & 88.2        & 68.2        & 30.6        & 81.1        & 49.6        & 61.0        & 77.8        & 66.1        & 75.1        & 29.0        & 66.0\\
%MCOF~\cite{MCOF}                   & 87.0        & 78.4        & 29.4        & 68.0        & 44.0        & 67.3        & 80.3        & 74.1        & 82.2        & 21.1        & 70.7\\
%Zeng et al.~\cite{ZENG}            & 90.0        & 77.4        & 37.5        & 80.7        & 61.6        & 67.9        & 81.8        & 69.0        & 83.7        & 13.6        & 79.4\\
%FickleNet~\cite{FICKLE}             & 89.5        & 76.6        & 32.6        & 74.6        & 51.5        & 71.1        & 83.4        & 74.4        & 83.6        & 24.1        & 73.4\\
%Sub-Categories~\cite{SUB}           & 88.8        & 51.6        & 30.3        & 82.9        & 53.0        & 75.8        &\textbf{88.6}& 74.8        & 86.6        &\textbf{32.4}& 79.9\\ \hline
PuzzleCAM                           & 88.6        &\underline{79.2}& 43.7        &\underline{89.0}& 61.8        & 72.1        & 83.3        & 76.0        &\underline{92.0}& 29.5        & \underline{86.0}\\
CLIMS                               & 89.8        & 71.2        &\underline{45.4}& 81.7        &\underline{70.2}& 67.6        & 84.0        & 75.7        & 90.0        & 20.3        & 84.2\\
PMM                                 & 89.5        & 76.8        & 43.9        & 88.1        & 65.8        & 76.0        &\underline{84.2}&\underline{78.0}& 91.2        & 30.6        & 84.3 \\
DRS                                 &\underline{90.1}& 78.9        & 45.3        & 85.8        & 68.4        &\underline{80.8}& 83.8        & 77.4        & 90.6        &\underline{31.5}& 84.0\\ \hline
ISLE-2 (Ours)             &90.5&\textbf{79.6}&\textbf{45.4}&\textbf{89.0}& 70.0        & 72.1        & 84.0        & 76.1        &\textbf{92.2}& 30.9        & \textbf{86.1}\\
ISLE (Ours)             &\textbf{91.0}&\textbf{79.6}&\textbf{45.4}&\textbf{89.0}& 69.7        &\textbf{81.2}&\textbf{84.2}&\textbf{78.0}&\textbf{92.2}&\textbf{31.6}& \textbf{86.1} \\ \hline
\end{tabular}}%}

\end{table*}

\begin{table*}[ht]
\centering
%\resizebox{\textwidth}{!}{

\caption{Semantic segmentation mIoU performance on the remaining 10 classes of the VOC2012 \textbf{training dataset} for the final pseudo-labels. \label{tab4}}
{\begin{tabular}{lcccccccccccccccccccccc}

\hline
Method                       & table          & dog           & horse       & motor        & person      & plant       & sheep       & sofa        & train       & tv          & mIoU\\ \hline
%AffinityNet~\cite{AFF}               & 40.2           & 80.4          & 62.0        & 70.4         &\textbf{73.7}& 42.5        & 70.7        & 42.6        & 68.1        & 51.6        & 61.7\\
%MCOF~\cite{MCOF}            & 28.2           & 73.2          & 71.5        & 67.2         & 53.0        & 47.7        & 74.5        & 32.4        & 71.0        & 45.8        & 60.3\\
%Zeng et al.~\cite{ZENG}     & 23.3           & 78.0          & 75.3        & 71.4         & 68.1        & 35.2        & 78.2        & 32.5        & 75.5& 48.0        & 63.3\\
%FickleNet~\cite{FICKLE}      & 47.4           & 78.2          & 74.0        & 68.8         &\textbf{73.2}& 47.8        & 79.9        & 37.0        & 57.3        &\textbf{64.6}& 64.9\\
%Sub-Categories~\cite{SUB}    & 53.8           & 82.3          & 78.5        & 70.4         & 71.2        & 40.2        & 78.3        & 42.9        & 66.8        & 58.8        & 66.1\\\hline
PuzzleCAM                    & 44.0           & \underline{91.6} &\underline{83.1}& \underline{80.1}& 42.8        &\underline{68.9}&\underline{92.6}& 53.4        & 64.8        & 42.6        & 69.7\\
CLIMS                        &\underline{57.8}   & 86.9          & 80.9        & 80.8         &\underline{72.7}& 48.4        & 90.3        &\underline{56.5}& 68.1        &\underline{58.4}& 70.5 \\
PMM                          & 48.5           & 89.3          & 82.0        & 79.0         & 61.4        & 66.5        & 89.9        & 54.4        & 66.4        & 38.6        & 70.7 \\
DRS                          & 41.2           & 88.7          & 80.0        & 79.8         & 65.4        & 62.6        & 89.9        & 55.0        &\underline{77.0}& 41.3        &\underline{71.3}\\ \hline
ISLE-2      & \textbf{52.5}           &\textbf{92.0}  &\textbf{86.0}&\textbf{80.9}& 71.8        & 68.5        &\textbf{92.7}& 56.2        & 68.1        & 54.1        & 73.3\\
ISLE      & 51.1           &91.9  &85.9&80.8& 72.3        & 68.5        &\textbf{92.7}&\textbf{56.9}&\textbf{77.1}& 52.8        & \textbf{74.2} \\ \hline
\end{tabular}}%}

\end{table*}

\subsection{Semantic Segmentation Performance on VOC2012}

Next, we compare our ensemble with its components consisting of recent works using image-level supervision Table~\ref{tab2}.
We trained all pseudo-labels with the same DeepLabV3+ model for comparability using a ResNet50 backbone.
We notice that the ensemble outperforms its component by a margin of at least $2\%$, although the individual components do not show this amount of variance between them. ISLE-2 is the ensemble of just PuzzleCAM and CLIMS, and ISLE is the ensemble of all four methods.
DRS is the best performing of its component, and the ISLE reaches a $2\%$ higher mIoU score.

Here, we present a more comprehensive analysis by providing a class-wise mIoU breakdown for all classes in the VOC2012 training dataset and a discussion.
On the one hand, we see in Table~\ref{tab3} and Table~\ref{tab4}  that the difference in the average mIoU score between our four component methods the relatively small, with the lowest scoring PuzzleCAM reaching 69.7\% and the highest scoring DRS at 71.3\%.
On the other hand, the ensemble of all four methods reaches 74.1\%, and the ensemble of just PuzzleCAM and CLIMS 73.6\% achieves a significant gain compared to its components.
Although, we also notice that the gain from adding more image-level segmentation pseudo-labels to the ensemble shrinks over time and needs to be considered when choosing the component for ISLE, as the training set mIoU between ISLE and ISLE-2 differs by less than 1\%, while ISLE uses the double amount of components.

Let us take a closer look at the performance of the individual components on a class-wise basis.
PMM reaches the best score only in two classes and an average $0.77\%$ improvement in those classes.
Although only reaching the highest average score in three classes, DRS provides an average $6.04\%$ improvement in those three classes.
CLIMS is the best in seven classes and achieves an average improvement of $6.04\%$ as well.
Whereas PuzzleCAM is the lowest average scoring method but reaches high scores in eight classes but improves them only by $2.62\%$ on average.
Therefore, we conclude that CLIMS and PuzzleCAM contribute the most and PMM the least to the ensemble.
Hence, we also evaluated the combination of only CLIMS and PuzzleCAM to see how much improvement we gain while combining the minimum amount of pseudo-label sets. We called this version ISLE-2.
We notice that most high scores translated to the ensemble, with only minor losses in some classes, most probably due to overlap with other classes.

%%%%%%%%%%%%%%%%%%%%%%%%%%%%%%%%%%%%%%%%%%%%%%%%%%%%%%%%%%%%%%%%%%%%%%%%%%%%%%%%%%%%%%%%
\begin{figure*}[ht]
\centering
\includegraphics[width=0.8\columnwidth]{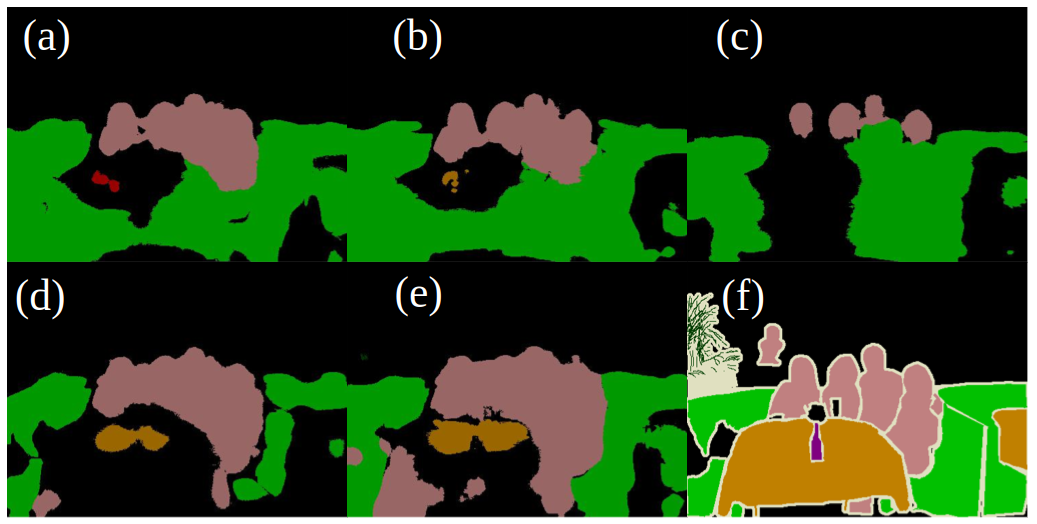} %width=80mm,scale=0.80 width=\linewidth
\caption{Pseudo labels from (a) DRS, (b) PMM, (c) PuzzleCAM, (d) CLIMS, (e)
ISLE (Ours), (f) Ground truth \label{exam2}.}\end{figure*}
%%%%%%%%%%%%%%%%%%%%%%%%%%%%%%%%%%%%%%%%%%%%%%%%%%%%%%%%%%%%%%%%%%%%%%%%%%%%%%%%%%%%%%%%

Fig.~\ref{exam2} presents one example of our experimental results.
(a) shows the pseudo labels of DRS, (b) of PMM, (c) of PuzzleCAM, (d) of CLIMS, (e) of our ISLE, and (f) the Ground truth.
We observe that DRS does not recognize the table in the image but over-detects the couch.
PMM has a few pixels detected as table but also detects fewer person pixels.
PuzzleCAM struggles even more with couch over-prediction and person under-prediction.
The best result of the not combined images stems from CLIMs, which only correct couch predictions and the best person detection.
Finally, our ISLE expands person detection but also includes some over-predictions, for example in the bottom left.
Furthermore, ISLE has the best table detection and successfully expands couch detections.

\subsection{Complexity Analysis}

In this section, we will provide a complexity analysis of our code.
This will help to estimate the additional complexity that is connected with adding more components.

\textbf{Step 1}

Let $\{ Comp_1, Comp_2, ..., Comp_N \}$ be the list of Components. Each component takes as input a specific image $i$ from the list of all images $I$ and gives as output class activation maps $CAM_n^{i,c}$ for all classes $c$ in the dataset.
$$Comp_n(i) =  \sum_{c=0}^C CAM_n^{i,c}, 1 \leq n \leq N$$
As the components are not further defined by the framework, we can only summarize their complexity as follows:
$$O(Step1) = \sum_{n=0}^N O(Comp_n^{training}(i)) \times O(Comp_n^{inference}(i)) \times epochs_{n} \times 2 \times I$$

\textbf{Step 2}

Let $\{ Ref_1, Ref_2, ..., Ref_M \}$ be the list of all applied refinements. We assume that all refinements are applied to all components for ease of notation.
Then Step. 2 is defined as:
$$\widetilde{CAM_n^{i}} = Ref_1(CAM_n^{i}) \otimes Ref_2(CAM_n^{i}) \otimes ... \otimes Ref_M(CAM_n^{i}),$$
for any $n$ with $1 \leq n \leq N$ Again, we need to define the complexity of each refinement method as $O(Ref_m())$ as the refinements are not further defined by the framework:
$$O(Step2) = \sum_{m=0}^M O(Ref_m^{training}(i)) \times O(Ref_m^{inference}(i)) \times epochs_{n} \times 2 \times I \times N$$

\textbf{Step 3}

The merging of pseudo-labels is done after a class-wise evaluation for each $\widetilde{CAM_n^{i}}$ to determine which Component after refinement has the high score for each class $c$ with 
$1 \leq c \leq C$ :
$$AE(i) = \sum_{c=1}^C AE^c(i) = \sum_{c=1}^C best(\widetilde{CAM_n^{c,i}}),$$
for all $i$ in $I$
The refinement step and Class-Wise Ensemble are just linearly dependent on the number of Components: 
$$O(Step3) =  O(eval) + O(merger)  = I \times N \times C + I \times C $$

\textbf{Step 4}

The training of the DeepLabV3+ model is not different from any other WSSS pipeline:
$$O(Step.4) = O(\texttt{DeepLabV3+}^{training} \times epochs \times images)$$

Nonetheless, for the final deployment of ISLE, only the forward pass of DeepLabV3+ is necessary, independent of the number of components and refinements used: 

$$O(Deployment) = O(\texttt{DeepLabV3+}^{inference} \times images)$$

\section{Conclusion}

In this paper, we have proposed our ISLE framework, which combines the pseudo-labels of several image-level segmentation techniques on a class-wise basis to leverage the strong points of its different components.
The combined pseudo labels reach at least 2\% higher mIoU scores than
its components.
Most of those gains stem from bigger variances within particular classes, as we observed that different approaches have different strengths and weaknesses.
The ISLE framework combines any number of pseudo-labels to boost the quality of the pseudo-labels for final training.
We showed that the predictions generated by the model trained with the pseudo labels of ISLE achieve state-of-the-art performance on the VOC2012 dataset showing its effectiveness.
Our framework is open-source to ensure reproducible research and accessibility.
The source code is accessible at \url{https://github.com/ErikOstrowski/ISLE}.

%\section*{Acknowledgments}
%This work is part of the Moore4Medical project funded by the ECSEL Joint Undertaking under grant number H2020-ECSEL-2019-IA-876190.
\section*{Acknowledgments}
This work is part of the Moore4Medical project funded by the ECSEL Joint Undertaking under grant number H2020-ECSEL-2019-IA-876190.
This work was also supported in parts by the NYUAD’s Research Enhancement Fund (REF) Award on “eDLAuto: An Automated Framework for Energy-Efficient Embedded Deep Learning in Autonomous Systems”, and by the NYUAD Center for Artificial Intelligence and Robotics (CAIR), funded by Tamkeen under the NYUAD Research Institute Award CG010.

\bibliographystyle{ieeetr}
\bibliography{our_framework}

\begin{thebibliography}{10}

\bibitem{AD}
J.~Ren, H.~Gaber, and S.~S. Al~Jabar, ``Applying deep learning to autonomous
  vehicles: A survey,'' in {\em 2021 4th International Conference on Artificial
  Intelligence and Big Data (ICAIBD)}, pp.~247--252, IEEE, 2021.

\bibitem{feng2020deep}
D.~Feng, C.~Haase-Sch{\"u}tz, L.~Rosenbaum, H.~Hertlein, C.~Glaeser, F.~Timm,
  W.~Wiesbeck, and K.~Dietmayer, ``Deep multi-modal object detection and
  semantic segmentation for autonomous driving: Datasets, methods, and
  challenges,'' {\em IEEE Transactions on Intelligent Transportation Systems},
  vol.~22, no.~3, pp.~1341--1360, 2020.

\bibitem{diakogiannis2020resunet}
F.~I. Diakogiannis, F.~Waldner, P.~Caccetta, and C.~Wu, ``Resunet-a: A deep
  learning framework for semantic segmentation of remotely sensed data,'' {\em
  ISPRS Journal of Photogrammetry and Remote Sensing}, vol.~162, pp.~94--114,
  2020.

\bibitem{meenpal2019facial}
T.~Meenpal, A.~Balakrishnan, and A.~Verma, ``Facial mask detection using
  semantic segmentation,'' in {\em 2019 4th International Conference on
  Computing, Communications and Security (ICCCS)}, pp.~1--5, IEEE, 2019.

\bibitem{khan2015multi}
K.~Khan, M.~Mauro, and R.~Leonardi, ``Multi-class semantic segmentation of
  faces,'' in {\em 2015 IEEE International Conference on Image Processing
  (ICIP)}, pp.~827--831, IEEE, 2015.

\bibitem{milioto2018real}
A.~Milioto, P.~Lottes, and C.~Stachniss, ``Real-time semantic segmentation of
  crop and weed for precision agriculture robots leveraging background
  knowledge in cnns,'' in {\em 2018 IEEE international conference on robotics
  and automation (ICRA)}, pp.~2229--2235, IEEE, 2018.

\bibitem{barth2018data}
R.~Barth, J.~IJsselmuiden, J.~Hemming, and E.~J. Van~Henten, ``Data synthesis
  methods for semantic segmentation in agriculture: A capsicum annuum
  dataset,'' {\em Computers and electronics in agriculture}, vol.~144,
  pp.~284--296, 2018.

\bibitem{rehman2020deep}
A.~Rehman, S.~Naz, M.~I. Razzak, F.~Akram, and M.~Imran, ``A deep
  learning-based framework for automatic brain tumors classification using
  transfer learning,'' {\em Circuits, Systems, and Signal Processing}, vol.~39,
  no.~2, pp.~757--775, 2020.

\bibitem{zhao2017tracking}
Z.~Zhao, S.~Voros, Y.~Weng, F.~Chang, and R.~Li, ``Tracking-by-detection of
  surgical instruments in minimally invasive surgery via the convolutional
  neural network deep learning-based method,'' {\em Computer Assisted Surgery},
  vol.~22, no.~sup1, pp.~26--35, 2017.

\bibitem{CT}
M.~Cordts, M.~Omran, S.~Ramos, T.~Rehfeld, M.~Enzweiler, R.~Benenson,
  U.~Franke, S.~Roth, and B.~Schiele, ``The cityscapes dataset for semantic
  urban scene understanding,'' in {\em Proceedings of the IEEE conference on
  computer vision and pattern recognition}, pp.~3213--3223, 2016.

\bibitem{CAM}
B.~Zhou, A.~Khosla, A.~Lapedriza, A.~Oliva, and A.~Torralba, ``Learning deep
  features for discriminative localization,'' in {\em Proceedings of the IEEE
  conference on computer vision and pattern recognition}, pp.~2921--2929, 2016.

\bibitem{VOC}
M.~Everingham, L.~Van~Gool, C.~K. Williams, J.~Winn, and A.~Zisserman, ``The
  pascal visual object classes (voc) challenge,'' {\em International journal of
  computer vision}, vol.~88, no.~2, pp.~303--338, 2010.

\bibitem{AFF}
J.~Ahn and S.~Kwak, ``Learning pixel-level semantic affinity with image-level
  supervision for weakly supervised semantic segmentation,'' in {\em
  Proceedings of the IEEE conference on computer vision and pattern
  recognition}, pp.~4981--4990, 2018.

\bibitem{PUZZLE}
S.~Jo and I.-J. Yu, ``Puzzle-cam: Improved localization via matching partial
  and full features,'' in {\em 2021 IEEE International Conference on Image
  Processing (ICIP)}, pp.~639--643, IEEE, 2021.

\bibitem{CLIMS}
J.~Xie, X.~Hou, K.~Ye, and L.~Shen, ``Clims: Cross language image matching for
  weakly supervised semantic segmentation,'' in {\em Proceedings of the
  IEEE/CVF Conference on Computer Vision and Pattern Recognition},
  pp.~4483--4492, 2022.

\bibitem{PMM}
Y.~Li, Z.~Kuang, L.~Liu, Y.~Chen, and W.~Zhang, ``Pseudo-mask matters in
  weakly-supervised semantic segmentation,'' in {\em Proceedings of the
  IEEE/CVF International Conference on Computer Vision}, pp.~6964--6973, 2021.

\bibitem{DRS}
B.~Kim, S.~Han, and J.~Kim, ``Discriminative region suppression for
  weakly-supervised semantic segmentation,'' in {\em Proceedings of the AAAI
  Conference on Artificial Intelligence}, vol.~35, pp.~1754--1761, 2021.

\bibitem{CRF}
Z.-H. Yuan, T.~Lu, Y.~Wu, {\em et~al.}, ``Deep-dense conditional random fields
  for object co-segmentation.,'' in {\em IJCAI}, vol.~1, p.~2, 2017.

\end{thebibliography}
\end{document}